%% file: tech_report.tex
\documentclass{article}

\usepackage[preprint]{neurips_2025}
\usepackage[utf8]{inputenc} 
\usepackage[T1]{fontenc}    
\usepackage{hyperref}       

\usepackage{url}            
\usepackage{booktabs}       
\usepackage{amsfonts}       
\usepackage{nicefrac}       
\usepackage{microtype}      
\usepackage{xcolor}         

\usepackage{graphicx}
\usepackage{amsmath}
\usepackage{multirow}
\usepackage{multicol}
\usepackage{subcaption}
\usepackage{colortbl}
\usepackage{xcolor}

\usepackage{epigraph}
\usepackage{algpseudocode}
\usepackage{amsmath}
\usepackage{amssymb}
\usepackage{mathtools}
\usepackage{amsthm}
\usepackage{algorithm}
\theoremstyle{plain}
\newtheorem{theorem}{Theorem}[section]

\theoremstyle{definition}
\newtheorem{definition}[theorem]{Definition}

\theoremstyle{remark}

\usepackage{bm}

\usepackage{enumitem}
\setlist[itemize]{topsep=0pt,itemsep=0.2ex,parsep=0pt,partopsep=0pt,leftmargin=*,label=\textbullet}

\title{General Agentic Memory Via Deep Research}

\author{%
    B.Y. Yan$^{1}$, \ \ Chaofan Li$^{1}$,  \ \ Hongjin Qian$^{1,3}$, \ \ Shuqi Lu$^{1}$, \ \ Zheng Liu$^{1,4}$\thanks{Project lead} \\
  1. Beijing Academy of Artificial Intelligence \ \ 2. Renmin University of China \\ 3. Peking University \ \  4. Hong Kong Polytechnic University \\
  \texttt{\{chienqhj,zhengliu1026\}@gmail.com} \\ 
}

\begin{document}

\maketitle



\begin{abstract} 

Memory is critical for AI agents, yet the widely-adopted static memory, aiming to create readily available memory in advance, is inevitably subject to severe information loss. To address this limitation, we propose a novel framework called \textbf{general agentic memory (GAM)}. GAM follows the principle of “\textbf{just-in time (JIT) compilation}” where it focuses on creating optimized contexts for its client at runtime while keeping only simple but useful memory during the offline stage. To this end, GAM employs a duo-design with the following components. 1) \textbf{Memorizer}, which highlights key historical information using a lightweight memory, while maintaining complete historical information within a universal page-store. 2) \textbf{Researcher}, which retrieves and integrates useful information from the page-store for its online request guided by the pre-constructed memory. This design allows GAM to effectively leverage the agentic capabilities and test-time scalability of frontier large language models (LLMs), while also facilitating end-to-end performance optimization through reinforcement learning. In our experimental study, we demonstrate that GAM achieves substantial improvement on various memory-grounded task completion scenarios against existing memory systems. 

\end{abstract}

\section{Introduction} 

\epigraph{"\textit{Intelligence is not the ability to store information, but to know where to find it}."}{---Albert Einstein} 

AI agents become increasingly popular thanks to the rapid advancement of large language models (LLMs)~\cite{agent_llm}. Today, prototypes of AI agents are being deployed across many crucial domains, such as information seeking, software engineering, and scientific research, showcasing huge potential in improving the productivity of human society~\cite{agent_usage_1,agent_usage_2,agent_usage_3}. This widespread application, however, creates an urgent need to manage complex and rapidly expanding contexts, as AI agents must continuously integrate vast amounts of information generated by both their internal reasoning and external feedback~\cite{llmcontextwindowlimit}. To address this challenge, there has been growing interest in developing specialized memory systems that provide agents with essential contextual information to support downstream tasks~\cite{memorysurvey}. Most existing memory systems follow the principle of \textbf{Ahead-of-Time (AOT) Compilation}. Under this paradigm, substantial computation is performed during the offline stage to compress raw contexts as lightweight memory, while incoming requests are served primarily based on this pre-constructed memory~\cite{memagent,amem,mem0,memoryos}. Although widely adopted, this AOT-style approach suffers from critical limitations. 

$\boldsymbol{\star}$ \textit{\textbf{Memorization} is a form of data compression; thus, it is inevitably subject to \textbf{information loss}.} The precomputed memory, being a compressed representation of raw data, inevitably suffers from \textit{information loss}, making it difficult to satisfy the fine-grained information needs requested by client agents. In addition, such memory systems generally assume a \textit{static structure}, preventing them from flexibly adapting to ad-hoc or unforeseen requests that demand nuanced interpretation and integration of information. Finally, existing approaches often rely heavily on \textit{domain expertise and handcrafted heuristics} to determine how memory is constructed and organized, which further constrains generalization across domains and tasks of the AOT-style memory systems. 

$\boldsymbol{\star}$ \textit{\textbf{Search} is made as the core of memory, while {memorization} is conducted to enable effective search.} We argue that lossless memory can only be realized via searching over a database of the complete history, where the pre-computed memory is introduced to support such a search process. With this insight, we propose \textbf{General Agentic Memory (GAM)}, a novel memory framework for general AI agents following the principle of \textbf{Just-in-Time (JIT) Compilation}. During the offline stage, it creates a light memory for the crucial historical information while maintaining the complete historical information in the database. At runtime, it performs intensive computation, namely deep research, to generate a customized, high-utility context for its request based on the pre-constructed memory. 

\begin{figure}[t]
    \centering
    \includegraphics[width=\textwidth]{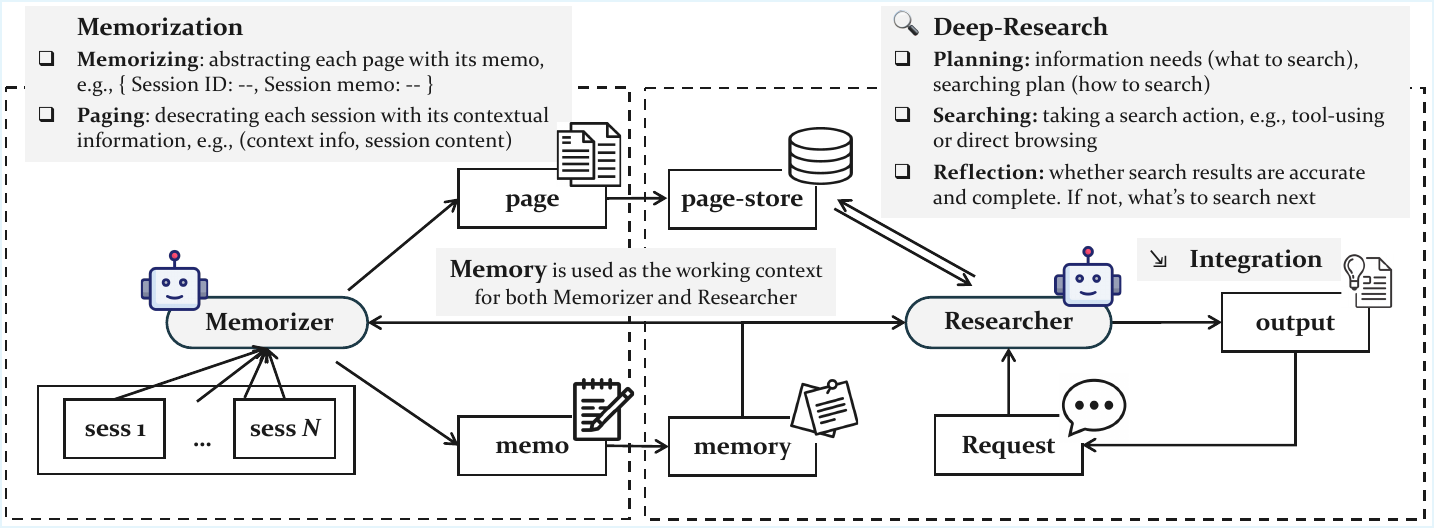} 
    \caption{Overview of GAM. The memorizer generates a light memory by for agent history and keeps the complete history in the page-store during the offline stage. The researcher performs deep-research to retrieve and integrate useful information for its request in the online service.} 
    \label{fig:frame} 
\end{figure}

$\boldsymbol{\star}$ \textit{\textbf{Dual-architecture}}. Based on the above JIT principle, GAM is realized based on a dual-agent framework with two fundamental roles: the Memorizer and the Researcher (Figure \ref{fig:frame}): 
\begin{itemize}
    \item The \textbf{Memorizer} receives the client's streaming history as a sequence of sessions, where it takes two actions: 1) it dynamically compresses the key historical information with a lightweight memory, and 2) it merges each session and its corresponding memory into a page and save all pages into a page-store, ensuring that the historical information is coherently and inclusively preserved. 
    \item The \textbf{Researcher} receives an online request from its client and performs deep research based on the pre-constructed memory to address the client’s needs. It iteratively analyzes information need and plans search actions, retrieves relevant information from the page-store, and reflects on the results until the gathered information fully satisfies the client’s request. 
\end{itemize} 
The above framework endows GAM with several key advantages. 1) \textit{High-fidelity and task-adaptability}, enabling the generation of concise yet highly informative memory tailored to downstream tasks. 2) \textit{Domain generalizability}, allowing GAM to operate effectively across general scenarios without relying on domain-specific expertise or handcrafted heuristics. 3) \textit{Optimizability}, harnessing advanced LLMs’ agentic capability and test-time scalability for performance optimization, while also facilitating continual improvement through reinforcement learning.

We evaluate GAM's performance through rigorous experimental studies. We jointly leverage the traditional memory benchmark LoCoMo~\cite{locomo}, together with popular long-context benchmarks such as HotpotQA~\cite{memagent}, RULER~\cite{ruler}, and NarrativeQA~\cite{kovcisky2018narrativeqa}. Across all these experiments, GAM consistently and significantly outperforms existing methods, demonstrating its strong ability to preserve fine-grained historical information and to optimize downstream task-completion performance for its clients. Our project is made publicly available to facilitate future research in this field\footnote{https://github.com/VectorSpaceLab/general-agentic-memory}. 


\section{Methodology}
\subsection{Definition} 
LLM agents often require long trajectories, comprising multi-step reasoning and tool using, to accomplish complex tasks, e.g., software engineering and deep research. In our work, we define each historical trajectory (history for short) as a sequence of temporally ordered units called sessions: $\mathrm{hist}: s_1, ..., s_T$. The rapidly growing history leads to several crucial challenges, including prohibitive computational costs, context window overflow, and performance degradation. To address these issues, a memory system is introduced to manage the information overload. Its primary objective is to extract useful yet concise information from the history, which is essential for the completion of the agent's task. That is to say, the memory system is to optimize the \textit{cost-effectiveness} of the agent’s task completion grounded on its produced context. This objective can be formulated as the following min–max optimization problem. 
\begin{definition} 
[\textbf{Memory}] A memory system produces the optimized context for an agent based on its task and history: $c^* \leftarrow \mathrm{Memory(task, history)}$, which is of the minimum size while optimizing the task completion performance: $c^* = \mathrm{argmin_{\mathcal{C}^*} ~ |c|}$, where $\mathcal{C}^* = \mathrm{argmax_{\mathcal{C}} ~Agent(task, context)}$. 
\end{definition} 

\subsection{General Agentic Memory}
The overall architecture of GAM, depicted in Figure \ref{fig:frame}, consists of two main modules: the memorizer and the researcher. Both modules are LLM-based agents, each with customized prompts\footnote{We include the detailed prompts of all functions in the appendix of the paper.}, working together to generate optimized memory that addresses requests from the client agent.

\subsubsection{Memorizer}
The {memorizer} is responsible for processing the agent’s trajectory during the offline stage, ensuring that it can be efficiently stored and effectively utilized. Each memorization step is triggered by the arrival of a new session ($s_i$), where two operations are performed. 1. \textit{Memorizing}, which produces memo ($\mu_i$) as a concise and well-structured snapshot of the new session. The memo is generated based on both the new session and the existing memory ($m_i$), highlighting its crucial information for the entire trajectory. The memory is therefore incrementally updated with the addition of the memo: 
\begin{equation}
    \mathrm{Memorizer.memorize}(s_i,m_i) \rightarrow {\mu}_i; ~ m_i+\{\mu_i\} \rightarrow m_{i+1}.
\end{equation}
2. \textit{Paging}, which creates pages to maintain the complete information of the agent’s trajectory. It begins by generating a header for the new session, which contains crucial contextual information from its preceding trajectory. The header is then used to decorate the session, forming a new page that is subsequently added to the page-store ($p$): 
\begin{equation}
    \mathrm{Memorizer.page}(s_i,m_i) \rightarrow h_i; ~ \{\mathrm{header}: h_i, ~ \mathrm{content}: s_i\} \rightarrow p_i; ~ p.\mathrm{append}(p_i).
\end{equation}
This process shares the same principle of BGE landmark retrieval~\cite{luo2024landmark} and Anthropic contextual retrieval~\cite{anthropic2023context}, which preserve the consistency of page semantics, ensuring that they can be accurately retrieved in subsequent stages. 

\subsubsection{Researcher}
The researcher is to address the client's request by retrieving and integrating useful information from the page-store. The process is iteratively conducted with three operations. 1) \textit{Planning}, which performs a chain-of-thought reasoning based on the existing memory to analyze the underlying information needed by request ($r$). Based on this initial reasoning result, it further generates concrete search plans according to the provided search toolkit ($\mathcal{T}$): 
\begin{equation}
    \mathrm{Researcher.plan}(r, m_i, \mathcal{T}) \rightarrow \{ \mathrm{tool}: t; ~ \mathrm{parameter}: \rho_t\}_{t \in \mathcal{T}}. 
\end{equation}
In our implementation, we offer three available tools for the researcher: an embedding model for vector search, a BM25 retriever for keyword-based search, and an ID-based retriever for direct page exploration. 2) \textit{Searching}. Upon obtaining the search plan, the researcher executes each search action in parallel, retrieving relevant pages ($p_t$) from the page-store. The researcher then integrates the information from the union of the retrieved pages together with the last integration result ($\mathcal{I}$) for the request ($r$), leading to an updated temporal integration result: 
\begin{equation}
    \text{For each $t$:}~ t(\rho_t)\rightarrow p_t; ~ \mathrm{Researcher.integrate}(\bigcup\nolimits_{t \in \mathcal{T}}p_t, \mathcal{I}, r) \rightarrow \mathcal{I}.  
\end{equation}
3) \textit{Reflection}. The researcher performs a reflection on whether the needed information in the request ($r$) has been fully collected by the integration result $\mathcal{I}$ using a binary indicator ($y$). If no, it further analyzes for the missing information, leading to a new request $r'$ to drive another round of deep research. If yes, the research process will be concluded by returning the integration result: 
\begin{equation}
    \mathrm{Researcher.reflect}(\mathcal{I}, r) \rightarrow y, r'; ~ \text{if $y$ = No, } \mathrm{Researcher}(r',\mathcal{I}); ~ \text{if $y$ = Yes, return $\mathcal{I}$}. 
\end{equation} 
Finally, the integrated result, along with the original information extracted from the associated pages, is returned to the client as the optimized context for its downstream task completion.

\subsubsection{Optimization}
A unified end-to-end performance optimization framework is introduced for GAM.
Suppose a training dataset $\mathcal{D}=\{(\mathrm{task}, \mathrm{hist})\}$ is given, the system creates the memory and page-store as: $\mathrm{M}, \mathrm{P} \leftarrow \mathrm{Memorizer}(\mathrm{hist})$, and then generates a candidate context for the task via: $c \leftarrow \mathrm{Researcher}(\mathrm{task},\mathrm{M},\mathrm{P})$. Using this candidate context, the client samples an answer ($\mathrm{ans}$), whose quality is measured by the reward function $\Gamma(\cdot)$. Thus, the expected reward is derived as: 
\begin{equation}
    \mathcal{R} = \mathbb{E}_{\mathrm{task,hist}\sim\mathcal{D}} ~ \mathbb{E}_{\mathrm{M}, \mathrm{P} \sim \mathrm{Memorizer} (\mathrm{hist})} ~
    \mathbb{E}_{c \sim \mathrm{Researcher}(\mathrm{task},\mathrm{M},\mathrm{P})} ~
    \mathbb{E}_{\mathrm{ans} \sim \mathrm{Client}(c,\mathrm{task})} ~
    \Gamma(\mathrm{ans}). 
\end{equation} 
When focusing on optimizing GAM's performance, the memorizer and the researcher are learned via reinforcement, while the client is excluded from the learning process. Without loss of generality, the policy gradients for the memorizer and researcher are given by: 
\begin{equation}
\begin{gathered}
    \nabla_{\theta_m} = \mathbb{E}_{\mathrm{task,hist}\sim\mathcal{D}} ~  (\Gamma(\mathrm{ans})-\bar{\Gamma}_{m}) \nabla_{\theta_m} \log \pi_m(\mathrm{M,P|hist}), \\
    \nabla_{\theta_r} = \mathbb{E}_{\mathrm{task,hist}\sim\mathcal{D}} ~ (\Gamma(\mathrm{ans})-\bar{\Gamma}_{r}) \nabla_{\theta_r} \log \pi_r(\mathrm{c}|\mathrm{task},\mathrm{M},\mathrm{P}).
\end{gathered}
\end{equation}
Here, $\theta_m$ and $\theta_r$ denote the model parameters of memorizer and researcher, respectively; $\bar{\Gamma}_{m}$ and $\bar{\Gamma}_{r}$ are the baseline answer rewards of the two modules; while $\theta_m(\cdot)$ and $\theta_r(\cdot)$ stand for the memorizer and researcher's generation likelihood. 

\section{Experiment} 

In this section, we conduct comprehensive experimental studies to evaluate the effectiveness of GAM. We focus on the investigation of the following three research questions. \textbf{RQ 1}: How does GAM perform compared with existing memory systems? \textbf{RQ 2}: How does GAM’s performance vary across different scenarios? \textbf{RQ 3}: How do key technical factors within GAM influence its performance? 


\subsection{Experiment Setting} 
\noindent \textbf{Datasets.} 
To rigorously evaluate the effectiveness of GAM, specifically 1) the memory’s ability to preserve historical information and 2) its ability to support downstream task completion, we employ the following benchmarks in our experimental studies. \textbf{1) LoCoMo}~\cite{locomo}. A widely used memory benchmark for conversational settings, designed to evaluate an agent’s ability to maintain and recall information across extended multi-session dialogues. We adopt its single-hop, multi-hop, temporal-reasoning, and open-domain tasks in our experiments. \textbf{2) HotpotQA}~\cite{hotpotqa}. A popular multi-hop question answering benchmark based on the Wikipedia corpus. We use the curated memory-evaluation dataset in MemAgent~\cite{memagent} that concatenates gold supporting documents with distracting passages. By varying the number of distractions, the dataset provides three versions with context lengths of 56K, 224K, and 448K tokens. \textbf{3) RULER}~\cite{ruler}. A popular long-context understanding benchmark with four types of evaluation tasks, including retrieval (Retri.), multi-hop tracing (MT), aggregation (AGG.), and question answering (QA). We use the 128K-token setting in our experiments. \textbf{4) NarrativeQA}~\cite{kovcisky2018narrativeqa}. A long-context question answering benchmark that provides an entire book or movie script as the input context for each sample. We randomly sample a subset of 300 questions for evaluation, whose average token length is 87K. 


\noindent \textbf{Baselines.}
We consider the following baselines in our experiments. 
\textbf{1) Memory-free methods}, including the \textit{brute-force long-LLM} (long-LLM for brevity) and retrieval-augmented generation (RAG). The long-LLM baseline attempts to process the entire input within the model’s context window. When the number of input tokens exceeds the maximum allowable context length $L_{max}$, the input is evenly partitioned into $N$ chunks of length $L_{max}$: $\{S_1, ..., S_N\}$, where the final score is reported as the maximum over all chunks: $\max \{ \mathrm{LLM}(S_1) ... \mathrm{LLM}(S_N) \}$. For the RAG baseline, the input is uniformly partitioned into segments of 2,048 tokens, and the top-5 retrieved segments are used to perform the downstream task. \textbf{2) Memory-based methods}, including A-Mem~\cite{amem}, Mem0~\cite{mem0}, MemoryOS~\cite{memoryos} and LightMem~\cite{lightmem}. These approaches construct specialized memory structures to store historical information, which can be utilized to address memory-related tasks at runtime.  

\noindent \textbf{Implementation Details.}
In our experiments, we adopt GPT-4o-mini and Qwen2.5-14B-Instruct~\cite{bai2023qwen} as the backbone models for both GAM and all baselines. Both LLMs offer a long-context window of 128K tokens. We use BGE-M3~\cite{chen2024bge} as the default dense retriever. For GAM’s detailed configuration, we set the maximum reflection depth to 3 and the maximum number of retrieved pages to 5. The input context is segmented into 2,048-token pages for stream processing in the memorization module.

\input{table/main_results_last}


\subsection{Main Results: Overall Effectiveness}
Table~\ref{tab:main} presents the main results of GAM and baselines on the experimental benchmarks, from which the following observations can be made. 
First, GAM consistently outperforms all baselines, including both memory-free and memory-based approaches, across every benchmark. Moreover, its advantage becomes particularly pronounced on benchmarks like HotpotQA and RULER, where tasks require multi-step retrieval and reasoning over information dispersed within the input context. For instance, GAM achieves over 90\% accuracy on the multi-hop tracing (MT) tasks in the RULER benchmark, which demand tracking variable values across multiple steps of assignment; in contrast, most baselines fail to achieve satisfactory performance under such complexity.
Finally, GAM maintains stable and competitive performance under varying input-context lengths, as reflected in the results on HotpotQA.
In summary, these experimental results preliminarily verify GAM’s overall effectiveness and its robustness to task complexity and growing context lengths. 


We obtain the following interesting things besides the main observations.
First, the performance of long-LLMs is under-expectation compared with the other methods, despite that it has adopted LLMs with a 128K context window, long enough to fully cover the input context in LoCoMo, HotpotQA-56K, and NarrativeQA. This suggests that {simply extending the context window is insufficient} to effectively address long-context challenges. This also aligns with the recently discussed phenomenon of context rot\footnote{https://research.trychroma.com/context-rot}, which indicates that the substantial distracting or irrelevant information within long contexts can severely degrade LLMs' performance. 
Second, direct applications of retrieval, i.e., RAG, exhibit highly variable performance across different scenarios. RAG improves performance over long-LLMs and the memory-based methods when the relevant information is explicitly presented, such as LoCoMo single-hop and RULER retrieval. However, it performs badly in HotpotQA, RULER multi-hop tracing, and RULER aggregation tasks, where relevant information is unobvious. In comparison, the memory-based methods show lower variance but remain constrained due to the loss of crucial details of the original context.
In contrast, GAM leverages memory to support effective retrieval of task-relevant information, enabling it to achieve substantially improved performance.


\subsection{Model's Impact}
Table~\ref{tab:mem_research_models} presents the performance of GAM on HotpotQA and NarrativeQA implemented with different LLMs. We apply Qwen-2.5 variants of different sizes (from 0.5B to 32B) and GPT-4o-mini as the backbones of the memorization and research module. As demonstrated by the experiment result, larger and stronger LLM-backbones for both memorizer and researcher result in consistent performance improvement, indicating that GAM can effectively leverage the increased LLM capacity to improve its memory quality. 
However, we also observe that the research module is considerably more sensitive to the LLM's scale than the memorization module. Notably, GAM maintains strong performance even when the memorizer is downsized and remains competitive with the smallest Qwen-2.5-0.5B model. In contrast, GAM’s overall performance deteriorates significantly when the research module’s backbone is reduced to 7B or smaller. This discrepancy reflects the distinct complexity of the two modules: the memorizer primarily extracts salient information from the input context, which is a relatively straightforward task, whereas the researcher must conduct iterative planning, searching, and reflection, which is much more complex and thus demands greater model capacity.

\input{table/different_models}



\begin{figure}[t]
    \centering
    
    \begin{subfigure}{\linewidth}
        \centering
        \includegraphics[width=\linewidth]{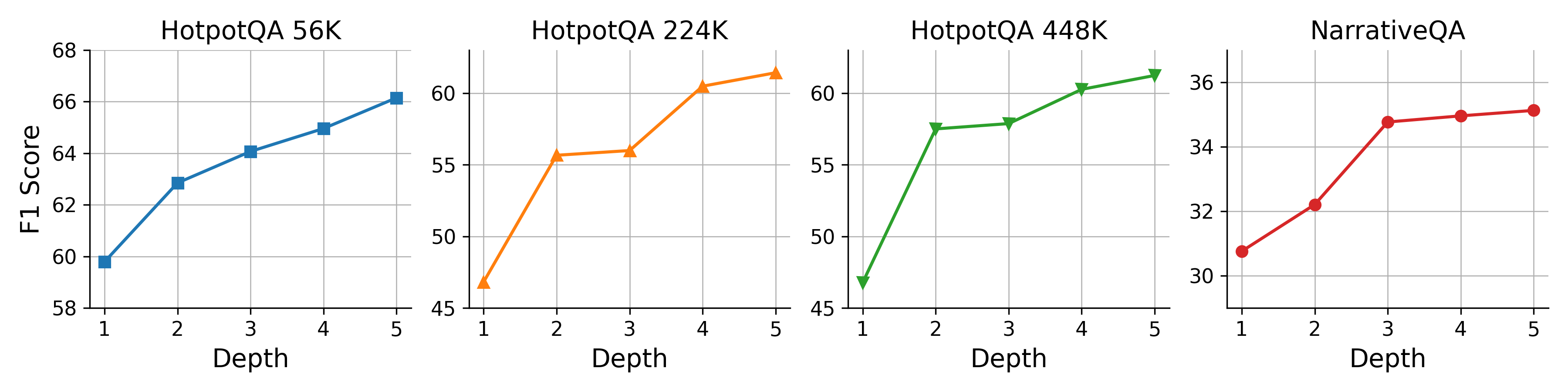}
        \caption{Impact of maximum reflection depth.}

        \label{fig:scaling:a}
    \end{subfigure}
    
    \vspace{0.4cm}  
    
    \begin{subfigure}{\linewidth}
        \centering
        \includegraphics[width=\linewidth]{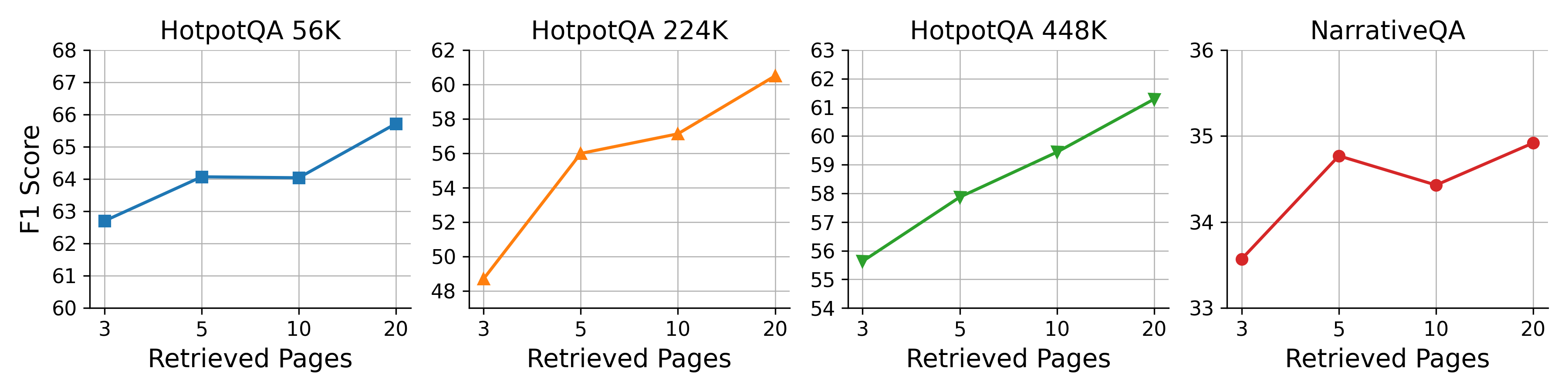}
        \caption{Impact of the amount of retrieved pages.}  
        \label{fig:scaling:b}
    \end{subfigure}
    
    \vspace{0.4cm}
    
    
    \caption{Impact of increasing test-time computation in reflection (top) and retrieval (bottom).}
    \label{fig:test-time}
\end{figure}

\input{table/ablation_last}

\input{table/integrate_results_last}




\subsection{Increasing Test-Time Computation}
As shown in Figure \ref{fig:test-time}, we investigate the impact of increasing test-time computation from two perspectives: 1) the depth of reflection and 2) the amount of retrieved pages. 
First, we vary the maximum reflection depth from 1 to 5 (3 by default), allowing GAM to perform additional research steps when necessary. Note that GAM autonomously determines the actual number of reflections and does not always reach the maximum step. This increased flexibility enables GAM to collect more information from the page-store, thus yielding consistent performance improvements across all datasets. However, the marginal gains gradually diminish, as many tasks do not require deep multi-step reasoning.
Second, we increase the number of retrieved pages from 3 to 20 (5 by default), enabling GAM to browse more pages in each step of research. The increase in retrieval results also leads to consistent performance improvements. 
Overall, both forms of increased test-time computation result in steady performance gains, which demonstrates GAM’s ability to benefit from test-time scaling, an advantage that baseline methods lack due to their fixed workflows.

\subsection{Detailed Factors' Analysis} 
We perform ablation studies to analyze other detailed influential factors, including searching tools, formation of GAM, and output formats.

First, we examine the impact of each searching tool and its combinations. As shown in Table~\ref{tab:ablation}, combining any two of the searching tools yields better performance than using each single tool alone, and the joint use of all three tools (i.e., GAM with the default setting) achieves the best performance. This observation validates the effectiveness of the search tools. Moreover, employing multiple tools enables broader exploration of the page-store, leading to better coverage of relevant information and, consequently, improved performance.

Second, we evaluate GAM’s performance when each module is used in isolation, namely 1) research without memory, and 2) memory without research. According to the experiment result in Table~\ref{tab:ablation}, using the research module alone leads to a substantial performance drop compared with the complete GAM system, highlighting the crucial role of memory in supporting effective exploration of relevant information. Using the memory module alone results in even worse performance, indicating that the pre-computed memory is prone to severe information loss. This observation further echoes our previous conclusion that the pre-constructed memory used in traditional ahead-of-time paradigms is far more limited than the just-in-time approach adopted by GAM. 

Third, we explore the impact of different forms of output, including 1) the researcher’s integration result (default), 2) the integration result accompanied by the relevant pages that provided its source information, and 3) the integration result paired with extracted source snippets from those relevant pages. As shown in Table~\ref{tab:integration_variants}, using only the integration result already achieves highly competitive performance. However, augmenting it with source information from the relevant pages yields further improvements, as it helps mitigate the loss of fine-grained details that may occur during integration.

\input{table/efficiency_last}

\subsection{Efficiency}
To assess the working efficiency of GAM, we measure the average time consumption, including both offline memory construction and online serving, when processing HotpotQA tasks under the 56K, 224K, and 448K settings. As shown in Table~\ref{tab:efficiency}, GAM incurs a time cost comparable to Mem0 and MemoryOS, and is substantially faster than A-mem. All methods exhibit approximately linear growth in offline construction time as context length increases, while maintaining relatively stable online serving time. Overall, GAM delivers strong performance with competitive efficiency, offering the best cost-effectiveness among experimental approaches.

\section{Conclusion}
In this paper, we present a novel memory system called General Agentic Memory (GAM), which is developed under the just-in-time compilation principle. GAM employs a dual-framework comprising a memorizer and a researcher. During the offline stage, the memorizer extracts the key information for its incoming context with lightweight memory and preserve the complete information within a page-store. For each online request, the researcher performs deep-research over the page-store based on the pre-constructed memory, which generates concise yet informative memory to support the downstream task. We perform comprehensive empirical studies using a variety of popular memory and long-context benchmarks, whose result validates the effectiveness of GAM given its significant and consistent improvements over existing methods.

\bibliographystyle{unsrt}
\bibliography{references_tech}

\clearpage

\section*{Appendix}

\subsection*{Baseline Reproduction Details}
When reproducing the baseline methods on the LocoMo dataset, we found that the category labels used for A-mem, Mem0, and MemoryOS were incorrect. Based on the official LocoMo annotations, we corrected the corresponding category–label mapping.

\subsection*{Prompts}

\begin{figure}[htbp]
    \centering
    \includegraphics[width=\textwidth]{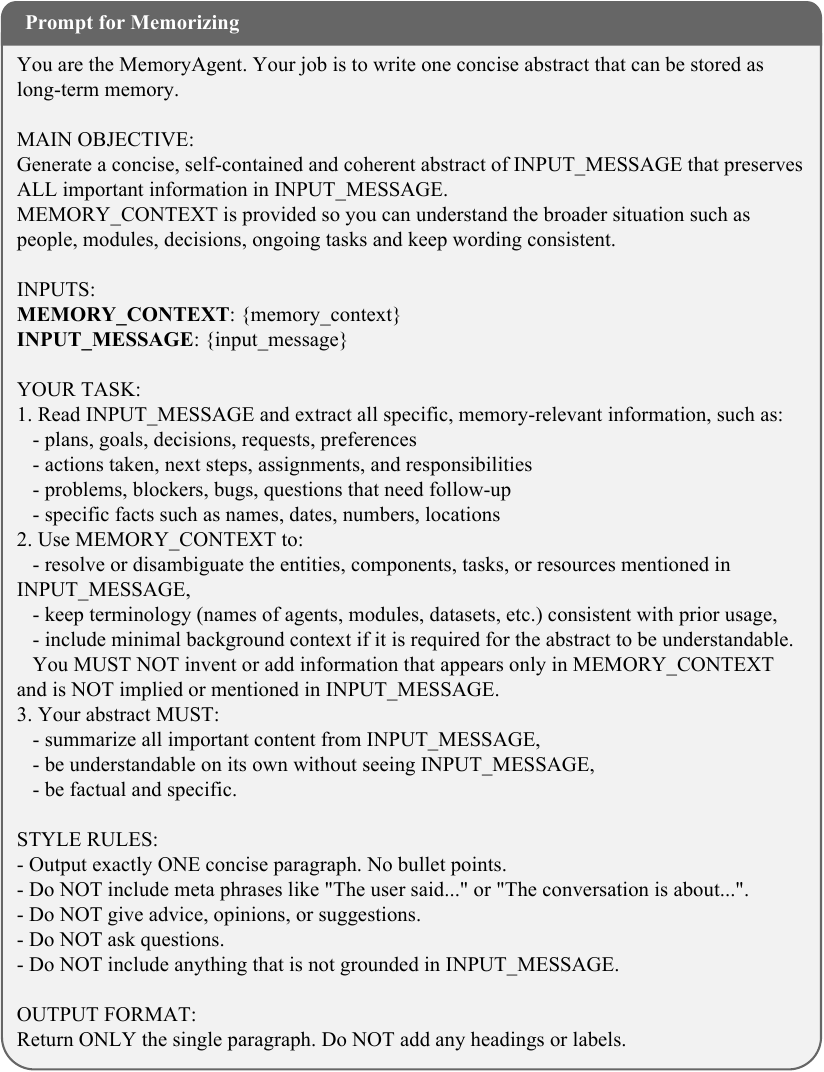} 
    \label{fig:memorize_prompt} 
\end{figure}


\begin{figure}[htbp]
    \centering
    \includegraphics[width=\textwidth]{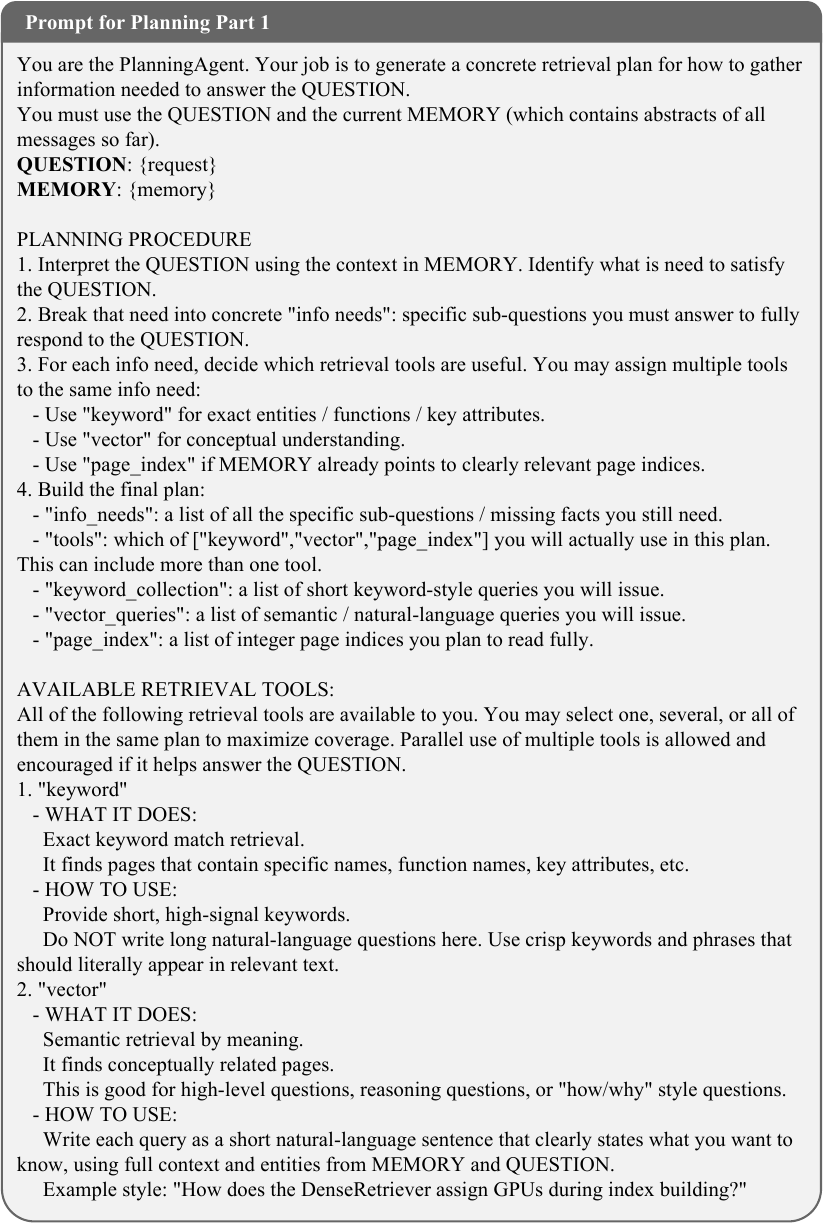} 
    \label{fig:planning_prompt} 
\end{figure}

\begin{figure}[htbp]
    \centering
    \includegraphics[width=\textwidth]{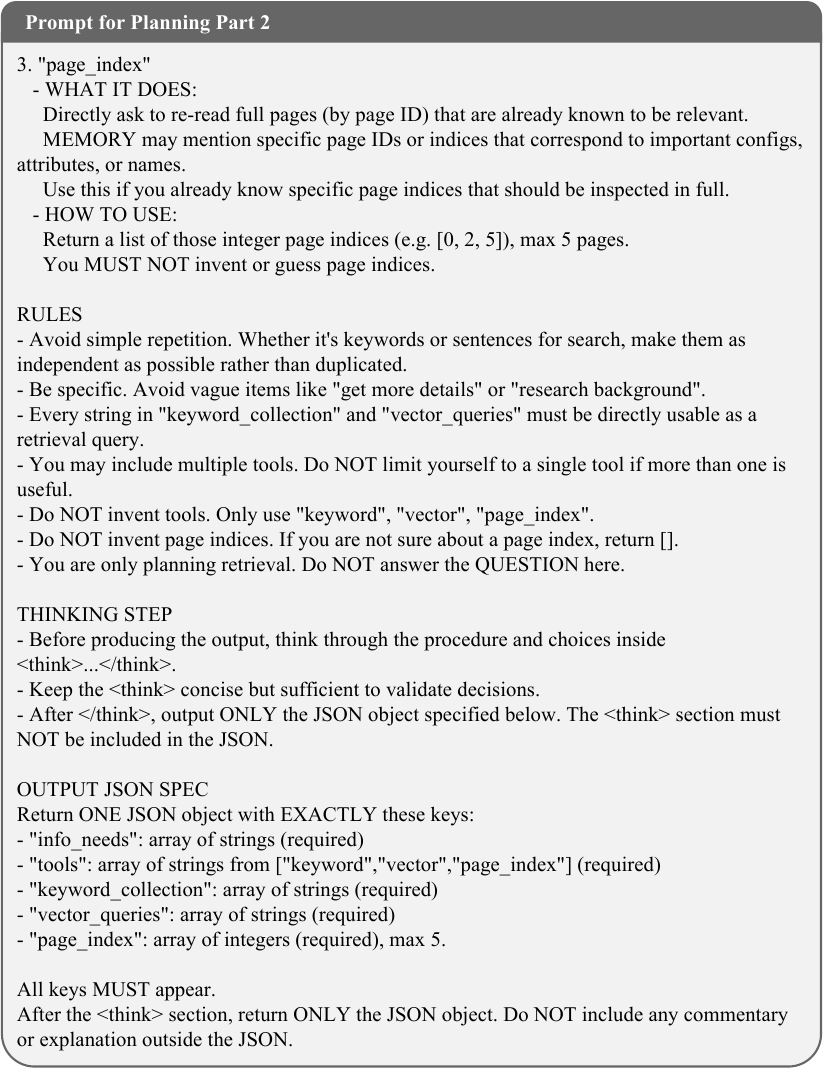} 
    \label{fig:planning_prompt} 
\end{figure}


\begin{figure}[htbp]
    \centering
    \includegraphics[width=\textwidth]{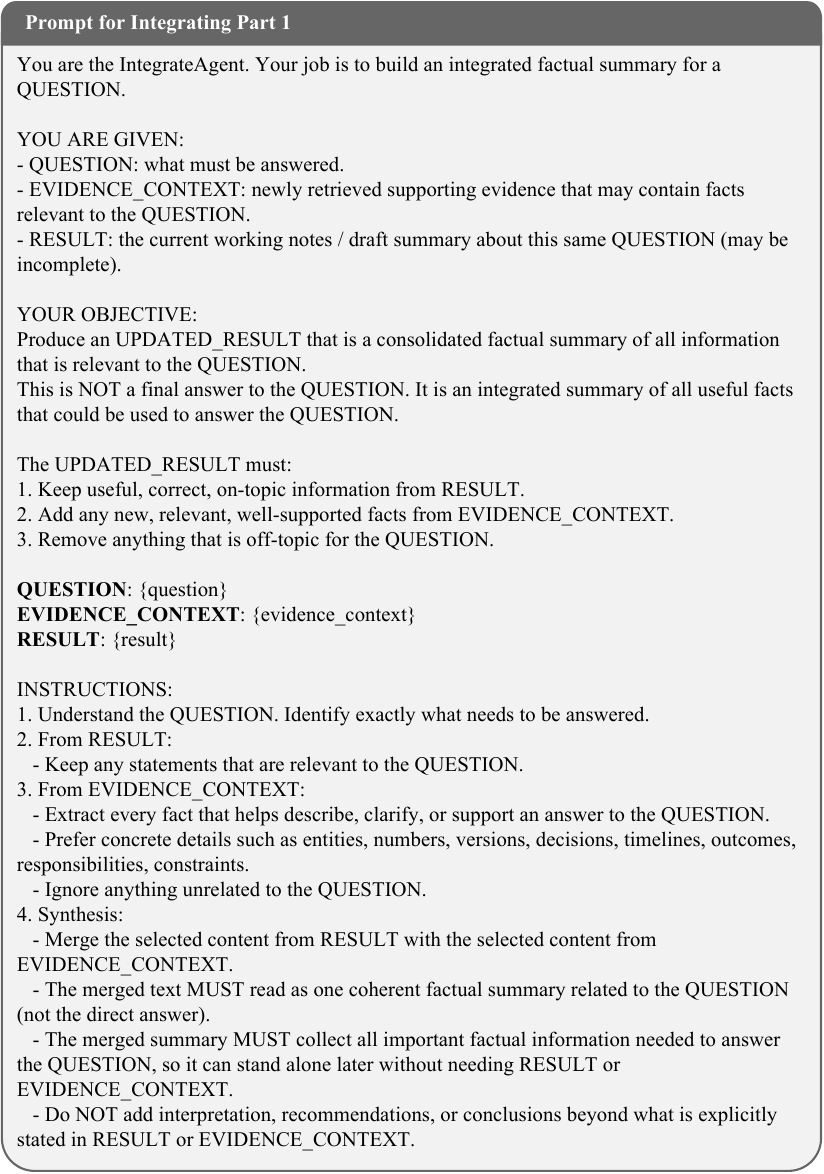} 
    \label{fig:integrate_prompt} 
\end{figure}

\begin{figure}[htbp]
    \centering
    \includegraphics[width=\textwidth]{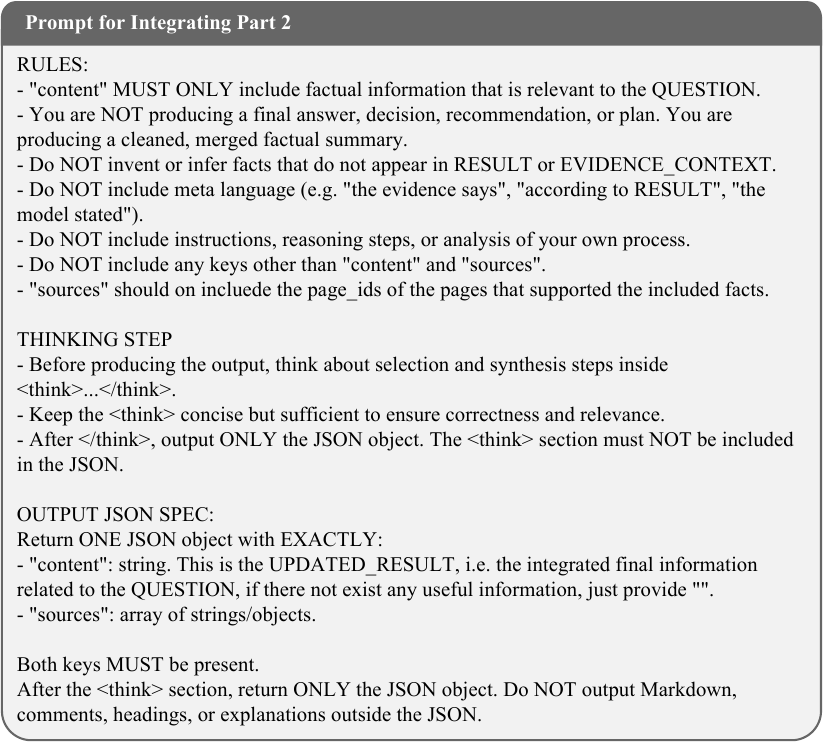} 
    \label{fig:integrate_prompt} 
\end{figure}

\begin{figure}[htbp]
    \centering
    \includegraphics[width=\textwidth]{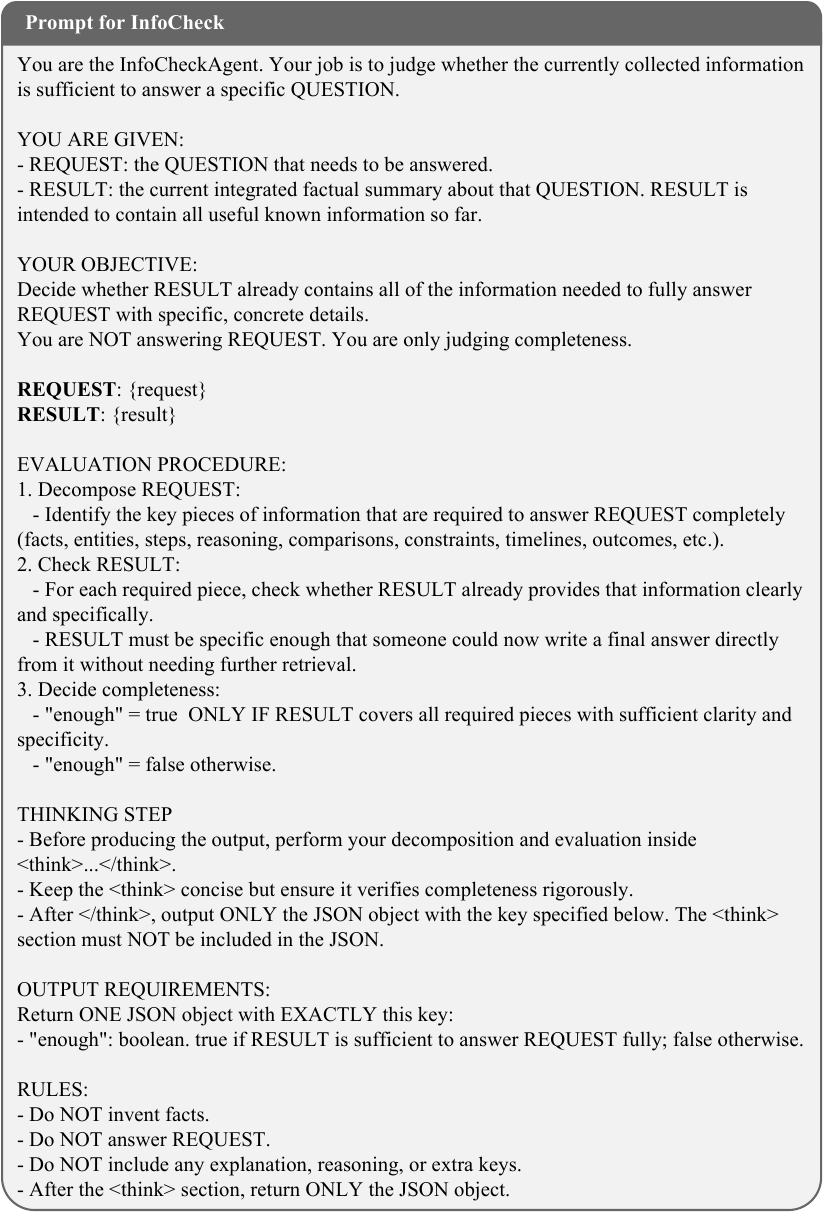} 
    \label{fig:integrate_prompt} 
\end{figure}

\begin{figure}[htbp]
    \centering
    \includegraphics[width=\textwidth]{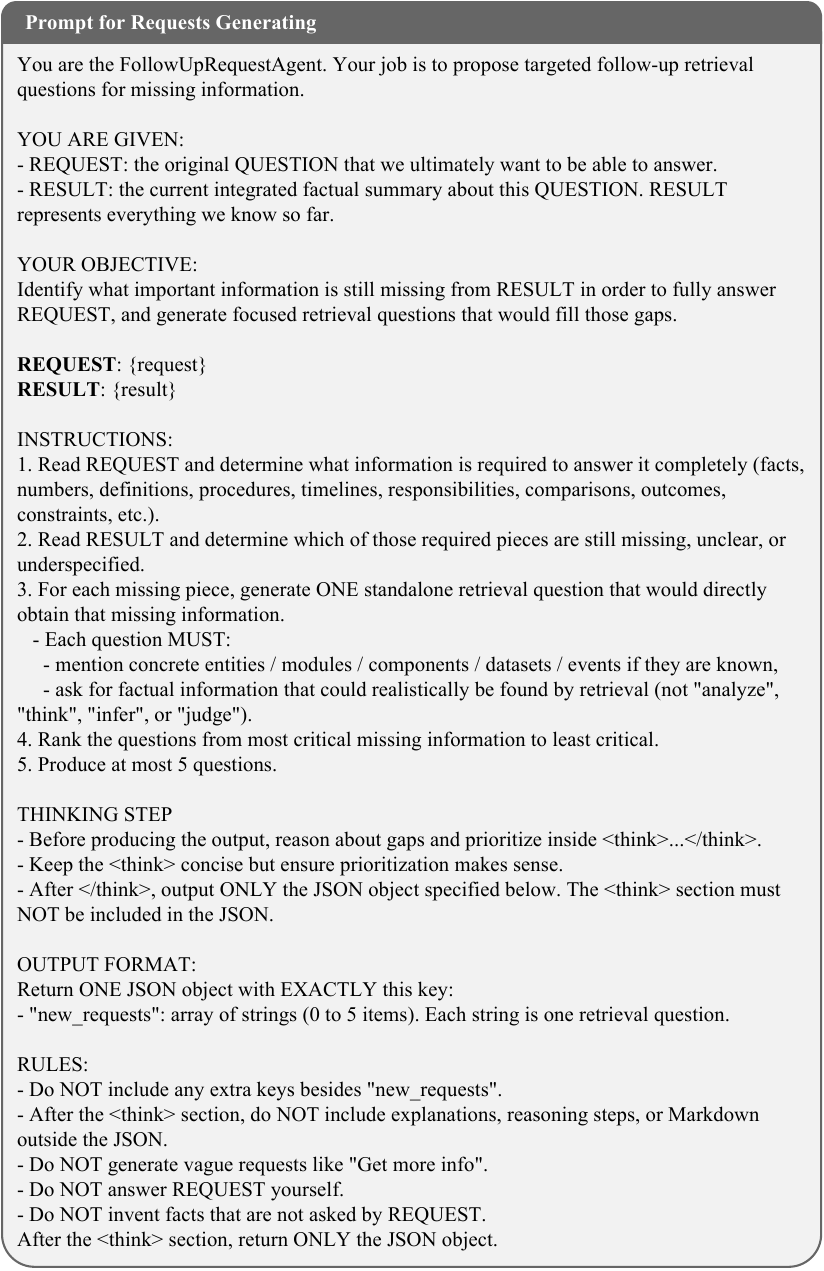} 
    \label{fig:integrate_prompt} 
\end{figure}

\end{document}

%% file: table/main_results_last.tex
\begin{table*}[t]
\centering
\small  
\caption{Results from GAM and baselines (memory-free and memory-based) on LoCoMo, HotpotQA, RULER, and NarrativeQA. Two LLMs, GPT-4o-mini and Qwen-2.5-14B, are used in experiment.} 
\label{tab:main}
\vspace{3pt}
\renewcommand{\arraystretch}{1.3}


\begin{subtable}{\textwidth}
\centering
\caption{Results on LoCoMo.}
\label{tab:main-locomo}
\begin{tabular}{cl|cccccccc}
\hline
\multicolumn{1}{c}{\textbf{Model}} &
\multicolumn{1}{l|}{\textbf{Method}} & \multicolumn{8}{c}{\textbf{LoCoMo}} \\
\cline{3-10}
& & \multicolumn{2}{c}{\textbf{Single Hop}} &
\multicolumn{2}{c}{\textbf{Multi Hop}} &
\multicolumn{2}{c}{\textbf{Temporal}} &
\multicolumn{2}{c}{\textbf{Open Domain}} \\
\cline{3-10}
& & \textbf{F1} & \textbf{BLEU-1} & \textbf{F1} & \textbf{BLEU-1} &
\textbf{F1} & \textbf{BLEU-1} & \textbf{F1} & \textbf{BLEU-1} \\
\hline


\multirow{7}{*}{\textbf{\rotatebox{90}{GPT-4o-mini}}}
& \textsc{long-LLM}         
& 46.68 & 37.54 & 29.23 & 22.76 & 25.97 & 19.42 & 16.87 & 13.70 \\

& \textsc{RAG}                
& \underline{52.45} & \underline{47.94} & 27.50 & 20.13 & 46.07 & 40.35 & 23.23 & 17.94 \\
\cline{2-10}

& \textsc{A-mem}              
& 44.65 & 37.06 & 27.02 & 20.09 & 45.85 & 36.67 & 12.14 & 12.00 \\

& \textsc{Mem0}               
& 47.65 & 38.72 & \underline{38.72} & \underline{27.13} & \underline{48.93} & \underline{40.51} & \underline{28.64} & \underline{21.58} \\

& \textsc{MemoryOS}           
& 48.62 & 42.99 & 35.27 & 25.22 & 41.15 & 30.76 & 20.02 & 16.52 \\

& \textsc{LightMem}           
& 41.79 & 37.83 & 29.78 & 24.80 & 43.71 & 39.72 & 16.89 & 13.92 \\
\cline{2-10}

& \textsc{GAM}                
& \textbf{57.75} & \textbf{52.10} & \textbf{42.29} & \textbf{34.44} &
  \textbf{59.45} & \textbf{53.11} & \textbf{33.30} & \textbf{26.97} \\ 
\hline


\multirow{7}{*}{\textbf{\rotatebox{90}{Qwen2.5 14b}}}
& \textsc{long-LLM}         
& 46.05 & 39.56 & 32.08 & 24.46 & 30.51 & 24.45 & 14.89 & 11.41 \\

& \textsc{RAG}                
& \underline{47.87} & \underline{42.79} & 26.38 & 19.54 & 30.78 & 25.97 & 14.16 & 10.52 \\
\cline{2-10}

& \textsc{A-mem}              
& 33.75 & 30.04 & 22.09 & 15.28 & 27.19 & 22.05 & 13.49 & 10.74 \\

& \textsc{Mem0}               
& 42.58 & 35.15 & 31.73 & 24.82 & 28.96 & 26.24 & 15.03 & 11.28 \\

& \textsc{MemoryOS}           
& 46.33 & 41.62 & \underline{38.19} & \underline{29.26} &
  \underline{32.24} & \underline{27.86} & \underline{20.27} & \underline{15.94} \\

& \textsc{LightMem}           
& 34.92 & 31.22 & 25.45 & 19.61 & 32.03 & 27.70 & 15.81 & 11.81 \\
\cline{2-10}

& \textsc{GAM}                
& \textbf{58.93} & \textbf{53.74} & \textbf{42.96} & \textbf{34.48} &
  \textbf{51.52} & \textbf{44.43} & \textbf{30.63} & \textbf{26.04} \\
\hline

\end{tabular}
\end{subtable}

\vspace{6pt}


\begin{subtable}{\textwidth}
\centering
\caption{Results on HotpotQA, RULER, and NarrativeQA.}
\label{tab:main-other}
\begin{tabular}{cl|ccc|cccc|c}
\hline
\multicolumn{1}{c}{\textbf{Model}} &
\textbf{Method} &
\multicolumn{3}{c|}{\textbf{HotpotQA}} &
\multicolumn{4}{c|}{\textbf{RULER(128k)}} &
\textbf{NarrativeQA} \\
\cline{3-10}
& & \textbf{56K} & \textbf{224K} & \textbf{448K} & 
\textbf{Retri.} & \textbf{MT} & \textbf{AGG.} & \textbf{QA} & \multirow{2}{*}{F1}\\
& & \textbf{F1} & \textbf{F1} & \textbf{F1} &
\textbf{Acc.} & \textbf{Acc.} & \textbf{Acc.} & \textbf{Acc.} & \\
\hline


\multirow{7}{*}{\textbf{\rotatebox{90}{GPT-4o-mini}}}
& \textsc{long-LLM}         
& \underline{56.56} & \underline{54.29} & 53.92 &
  80.30 & \underline{60.60} & \underline{36.70} & \underline{61.60} &
  \underline{31.26} \\

& \textsc{RAG}                
& 52.71 & 51.84 & \underline{54.01} &
  \underline{94.25} & 0.00 & 35.50 & 55.90 &
  25.00 \\
\cline{2-10}

& \textsc{A-mem}              
& 33.90 & 30.22 & 31.37 &
  44.23 & 0.00 & 29.20 & 46.50 &
  27.07 \\

& \textsc{Mem0}               
& 32.58 & 31.74 & 27.41 &
  46.83 & 53.80 & 34.10 & 51.70 &
  29.16 \\

& \textsc{MemoryOS}           
& 26.47 & 23.10 & 24.16 &
  63.10 & 2.40 & 35.60 & 36.90 &
  26.70 \\

& \textsc{LightMem}           
& 40.93 & 35.28 & 30.02 &
  27.63 & 36.20 & 34.00 & 52.60 &
  17.51 \\
\cline{2-10}

& \textsc{GAM}                
& \textbf{63.22} & \textbf{64.56} & \textbf{59.81} &
  \textbf{97.70} & \textbf{93.20} & \textbf{42.50} & \textbf{72.50} &
  \textbf{36.86} \\ 
\hline


\multirow{7}{*}{\textbf{\rotatebox{90}{Qwen2.5 14b}}}
& \textsc{long-LLM}         
& 49.75 & \underline{46.82} & 43.17 &
  70.85 & \underline{80.00} & 15.40 & 45.60 &
  \underline{29.69} \\

& \textsc{RAG}                
& \underline{51.81} & 46.72 & \underline{48.36} &
  \underline{92.78} & 0.00 & 24.70 & 47.80 &
  18.29 \\
\cline{2-10}

& \textsc{A-mem}              
& 27.04 & 25.65 & 22.92 &
  39.73 & 0.00 & 25.80 & 40.20 &
  25.18 \\

& \textsc{Mem0}               
& 30.12 & 32.44 & 26.55 &
  43.03 & 41.20 & \underline{31.50} & 46.10 &
  27.80 \\

& \textsc{MemoryOS}           
& 24.58 & 30.25 & 23.13 &
  54.58 & 3.00 & 5.20 & 34.60 &
  23.45 \\

& \textsc{LightMem}           
& 37.30 & 27.72 & 28.25 &
  27.53 & 17.40 & 25.60 & \underline{53.00} &
  16.57 \\
\cline{2-10}

& \textsc{GAM}                
& \textbf{64.07} & \textbf{55.99} & \textbf{57.87} &
  \textbf{93.43} & \textbf{90.20} & \textbf{36.10} & \textbf{74.50} &
  \textbf{34.77} \\ 
\hline

\end{tabular}
\end{subtable}

\end{table*}

%% file: table/different_models.tex
\begin{table*}[t]
\centering
\caption{Model's impact on memorizer (left) and researcher (right), reflected by GAM’s performance.}
\label{tab:mem_research_models}
\vspace{3pt}

\begin{minipage}{0.48\textwidth}
\centering
\textbf{(a) Memorizer}\\[3pt]
\renewcommand{\arraystretch}{1.3}
\resizebox{\textwidth}{!}{%
\begin{tabular}{l|ccc|c|c}
\hline
\multirow{3}{*}{\textbf{Model}} &
\multicolumn{3}{c|}{\textbf{HotpotQA}} &
\textbf{NarrativeQA} &
\textbf{Avg} \\
\cline{2-6}
& \textbf{56K} & \textbf{224K} & \textbf{448K} & \multirow{2}{*}{\textbf{F1}} & \multirow{2}{*}{\textbf{F1}} \\
& \textbf{F1} & \textbf{F1} & \textbf{F1} & & \\
\hline
\textsc{Qwen2.5 0.5B}  & 56.46 & 55.96 & 53.33 & 29.55 & 48.83 \\
\textsc{Qwen2.5 3B}    & 58.05 & 56.52 & 55.50 & 32.10 & 50.54 \\
\textsc{Qwen2.5 7B}    & 59.06 & 58.34 & 56.17 & 32.53 & 51.53 \\
\textsc{Qwen2.5 14B}   & 64.07 & 55.99 & 57.87 & 34.77 & 53.18 \\
\textsc{Qwen2.5 32B}   & 63.05 & 59.75 & 56.26 & 34.94 & 53.50 \\
\textsc{GPT-4o mini}   & 64.77 & 59.29 & 57.25 & 34.87 & 54.05 \\
\hline
\end{tabular}%
}
\end{minipage}
\hfill
\begin{minipage}{0.48\textwidth}
\centering
\textbf{(b) Researcher}\\[3pt]
\renewcommand{\arraystretch}{1.3}
\resizebox{\textwidth}{!}{%
\begin{tabular}{l|ccc|c|c}
\hline
\multirow{3}{*}{\textbf{Model}} &
\multicolumn{3}{c|}{\textbf{HotpotQA}} &
\textbf{NarrativeQA} &
\textbf{Avg} \\
\cline{2-6}
& \textbf{56K} & \textbf{224K} & \textbf{448K} & \multirow{2}{*}{\textbf{F1}} & \multirow{2}{*}{\textbf{F1}} \\
& \textbf{F1} & \textbf{F1} & \textbf{F1} & & \\
\hline
\textsc{Qwen2.5 0.5B}  & 10.03 & 11.14 & 11.64 & 3.50 & 9.08 \\
\textsc{Qwen2.5 3B}    & 39.76 & 37.16 & 33.04 & 23.96 & 33.48 \\
\textsc{Qwen2.5 7B}    & 51.95 & 47.95 & 48.55 & 26.93 & 43.85 \\
\textsc{Qwen2.5 14B}   & 64.07 & 55.99 & 57.87 & 34.77 & 53.18 \\
\textsc{Qwen2.5 32B}   & 61.93 & 59.19 & 61.53 & 35.33 & 54.50 \\
\textsc{GPT-4o-mini}   & 62.06 & 62.97 & 61.54 & 35.24 & 55.45 \\
\hline
\end{tabular}%
}
\end{minipage}

\end{table*}

%% file: table/ablation_last.tex
\begin{table*}[htbp]
\centering
\small
\caption{Ablation study of detailed factors.}
\label{tab:ablation}
\vspace{3pt}
{%
\renewcommand{\arraystretch}{1.25}
\begin{tabular}{l|ccc|c|c}
\hline
\multirow{2}{*}{\textbf{Method}} &
\multicolumn{3}{c|}{\textbf{HotpotQA}} &
\textbf{NarrativeQA} &
\textbf{Avg} \\
\cline{2-6}
& \textbf{56K} & \textbf{224K} & \textbf{448K} & \multirow{2}{*}{\textbf{F1}} & \multirow{2}{*}{\textbf{F1}} \\
& \textbf{F1} & \textbf{F1} & \textbf{F1} & & \\
\hline

\textsc{GAM}       & 64.07 & 55.99 & 57.87 & 34.77 & 53.18 \\
\hline
\rowcolor{gray!20}
\multicolumn{6}{l}{\textbf{Tools}} \\

\quad \textsc{Only Page-id}          & 44.86 & 21.65 & 19.02 & 30.30 & 28.96 \\
\quad \textsc{Only Embedding}        & 39.59 & 32.71 & 26.67 & 30.25 & 32.31 \\
\quad \textsc{Only BM25}             & 59.24 & 52.29 & 51.52 & 31.50 & 48.64 \\
\quad \textsc{Embedding+Page-id}     & 47.25 & 34.78 & 28.43 & 33.41 & 35.97 \\
\quad \textsc{Embedding+BM25}        & 61.37 & 55.00 & 54.90 & 33.20 & 51.12 \\
\quad \textsc{BM25+Page-id}          & 63.57 & 55.38 & 55.62 & 32.05 & 51.66 \\

\hline

\rowcolor{gray!20}
\multicolumn{6}{l}{\textbf{Modules}} \\

\quad \textsc{Research without memory}    & 57.40 & 49.72 & 53.98 & 31.97 & 48.27 \\
\quad \textsc{Memory without research}     & 42.67 & 19.75 & 17.38 & 30.18 & 27.50 \\

\hline
\end{tabular}%
}
\end{table*}

%% file: table/integrate_results_last.tex
\begin{table*}[htbp]
\centering
\small
\setlength{\tabcolsep}{4.5pt}
\caption{Performance across different output formats.} 
\label{tab:integration_variants}
\vspace{3pt}
\renewcommand{\arraystretch}{1.3}
\begin{tabular}{l|l|ccc|c|c}
\hline
\multirow{2}{*}{\textbf{Model}} &
\multirow{2}{*}{\textbf{Metric}} &
\multicolumn{3}{c|}{\textbf{HotpotQA}} &
\textbf{NarrativeQA} &
\textbf{Avg} \\ 
\cline{3-7}
& & \textbf{56K} & \textbf{224K} & \textbf{448K} &  &  \\
\hline

\multirow{2}{*}{\textsc{Integration Only}} 
  & F1    & 64.07 & 55.99 & 57.87 & 34.77 & 53.18 \\
  & Tokens & 103.42 & 102.55 & 109.98 & 107.64 & 105.90 \\

\hline

\multirow{2}{*}{\textsc{Integration with Page}} 
  & F1    & 68.66 & 59.77 & 59.42 & 34.99 & 55.71 \\
  & Tokens & 1444.30 & 499.23 & 620.11 & 6955.62 & 2379.82 \\

\hline

\multirow{2}{*}{\textsc{Integration with Extraction}} 
  & F1    & 67.41 & 57.83 & 57.81 & 34.82 & 54.47 \\
  & Tokens & 220.78 & 227.57 & 230.47 & 244.20 & 230.76      \\

\hline
\end{tabular}%
\end{table*}

%% file: table/efficiency_last.tex
\begin{table*}[t]
\centering
\small
\caption{Efficiency analysis on HotpotQA}
\label{tab:efficiency}
\vspace{3pt}
\renewcommand{\arraystretch}{1.3}
\begin{tabular}{llccccc}
\hline
\textbf{Dataset} &
\textbf{Metric} &
\textbf{A-mem} &
\textbf{Mem0} &
\textbf{MemoryOS} &
\textbf{LightMem} &
\textbf{GAM} \\
\hline
\multirow{4}{*}{\textbf{HotpotQA 56k}} 
  & \textsc{Offline build} (s)           & 209.74 & 37.42  & 80.36 & 4.93  & 56.89 \\
  & \textsc{Online serve} (s)    & 0.52   & 0.15   & 0.44  & 0.20  & 12.43 \\
  & \textsc{Total} (s)             & 210.26 & 37.57  & 80.80 & 5.13  & 69.32 \\
  \cline{2-7}
  & \textsc{Answer quality} (F1)                  & 27.04  & 30.12  & 24.58 & 37.30 & 64.07 \\
\hline
\multirow{4}{*}{\textbf{HotpotQA 224k}} 
  & \textsc{Offline build} (s)          & 904.99  & 165.30  & 325.70 & 16.61    & 252.72 \\
  & \textsc{Online serve} (s)    & 0.48  & 0.17  & 0.55   & 0.25     & 16.65  \\
  & \textsc{Total} (s)                & 905.46  & 165.47  & 326.25 & 16.86    & 269.37 \\
  \cline{2-7}
  & \textsc{Answer quality} (F1)                & 25.65 & 32.44 & 30.25 & 27.72 & 55.99 \\
\hline
\multirow{4}{*}{\textbf{HotpotQA 448k}} 
  & \textsc{Offline build} (s)             & 1796.82  & 274.87  & 702.72 & 40.56  & 557.16 \\
  & \textsc{Online serve} (s)     & 0.47  & 0.18  & 0.46 & 0.21   & 18.49  \\
  & \textsc{Total} (s)                  & 1797.29 & 275.05 & 703.18  & 40.78  & 575.65 \\
  \cline{2-7}
  & \textsc{Answer quality} (F1)                 & 22.92 & 26.55 & 23.13 & 28.25 & 57.87 \\
\hline

\end{tabular}%
\end{table*}

%% file: tech_report.bbl
\begin{thebibliography}{10}

\bibitem{agent_llm}
Yuheng Cheng, Ceyao Zhang, Zhengwen Zhang, Xiangrui Meng, Sirui Hong, Wenhao Li, Zihao Wang, Zekai Wang, Feng Yin, Junhua Zhao, et~al.
\newblock Exploring large language model based intelligent agents: Definitions, methods, and prospects.
\newblock {\em arXiv preprint arXiv:2401.03428}, 2024.

\bibitem{agent_usage_1}
Xiang Deng, Yu~Gu, Boyuan Zheng, Shijie Chen, Sam Stevens, Boshi Wang, Huan Sun, and Yu~Su.
\newblock Mind2web: Towards a generalist agent for the web.
\newblock {\em Advances in Neural Information Processing Systems}, 36:28091--28114, 2023.

\bibitem{agent_usage_2}
Wei Tao, Yucheng Zhou, Yanlin Wang, Wenqiang Zhang, Hongyu Zhang, and Yu~Cheng.
\newblock Magis: Llm-based multi-agent framework for github issue resolution.
\newblock {\em Advances in Neural Information Processing Systems}, 37:51963--51993, 2024.

\bibitem{agent_usage_3}
Samuel Schmidgall, Yusheng Su, Ze~Wang, Ximeng Sun, Jialian Wu, Xiaodong Yu, Jiang Liu, Michael Moor, Zicheng Liu, and Emad Barsoum.
\newblock Agent laboratory: Using llm agents as research assistants.
\newblock {\em arXiv preprint arXiv:2501.04227}, 2025.

\bibitem{llmcontextwindowlimit}
Shouyuan Chen, Sherman Wong, Liangjian Chen, and Yuandong Tian.
\newblock Extending context window of large language models via positional interpolation.
\newblock {\em arXiv preprint arXiv:2306.15595}, 2023.

\bibitem{memorysurvey}
Zeyu Zhang, Quanyu Dai, Xiaohe Bo, Chen Ma, Rui Li, Xu~Chen, Jieming Zhu, Zhenhua Dong, and Ji-Rong Wen.
\newblock A survey on the memory mechanism of large language model-based agents.
\newblock {\em ACM Transactions on Information Systems}, 43(6):1--47, 2025.

\bibitem{memagent}
Hongli Yu, Tinghong Chen, Jiangtao Feng, Jiangjie Chen, Weinan Dai, Qiying Yu, Ya-Qin Zhang, Wei-Ying Ma, Jingjing Liu, Mingxuan Wang, et~al.
\newblock Memagent: Reshaping long-context llm with multi-conv rl-based memory agent.
\newblock {\em arXiv preprint arXiv:2507.02259}, 2025.

\bibitem{amem}
Wujiang Xu, Kai Mei, Hang Gao, Juntao Tan, Zujie Liang, and Yongfeng Zhang.
\newblock A-mem: Agentic memory for llm agents.
\newblock {\em arXiv preprint arXiv:2502.12110}, 2025.

\bibitem{mem0}
Prateek Chhikara, Dev Khant, Saket Aryan, Taranjeet Singh, and Deshraj Yadav.
\newblock Mem0: Building production-ready ai agents with scalable long-term memory.
\newblock {\em arXiv preprint arXiv:2504.19413}, 2025.

\bibitem{memoryos}
Jiazheng Kang, Mingming Ji, Zhe Zhao, and Ting Bai.
\newblock Memory os of ai agent.
\newblock {\em arXiv preprint arXiv:2506.06326}, 2025.

\bibitem{locomo}
Adyasha Maharana, Dong-Ho Lee, Sergey Tulyakov, Mohit Bansal, Francesco Barbieri, and Yuwei Fang.
\newblock Evaluating very long-term conversational memory of llm agents.
\newblock {\em arXiv preprint arXiv:2402.17753}, 2024.

\bibitem{ruler}
Cheng-Ping Hsieh, Simeng Sun, Samuel Kriman, Shantanu Acharya, Dima Rekesh, Fei Jia, Yang Zhang, and Boris Ginsburg.
\newblock Ruler: What's the real context size of your long-context language models?
\newblock {\em arXiv preprint arXiv:2404.06654}, 2024.

\bibitem{kovcisky2018narrativeqa}
Tom{\'a}{\v{s}} Ko{\v{c}}isk{\`y}, Jonathan Schwarz, Phil Blunsom, Chris Dyer, Karl~Moritz Hermann, G{\'a}bor Melis, and Edward Grefenstette.
\newblock The narrativeqa reading comprehension challenge.
\newblock {\em Transactions of the Association for Computational Linguistics}, 6:317--328, 2018.

\bibitem{luo2024landmark}
Kun Luo, Zheng Liu, Shitao Xiao, Tong Zhou, Yubo Chen, Jun Zhao, and Kang Liu.
\newblock Landmark embedding: a chunking-free embedding method for retrieval augmented long-context large language models.
\newblock In {\em Proceedings of the 62nd Annual Meeting of the Association for Computational Linguistics (Volume 1: Long Papers)}, pages 3268--3281, 2024.

\bibitem{anthropic2023context}
Anthropic.
\newblock Introducing contextual retrieval.
\newblock {\em https://www.anthropic.com/engineering/contextual-retrieval}, 2024.

\bibitem{hotpotqa}
Zhilin Yang, Peng Qi, Saizheng Zhang, Yoshua Bengio, William~W Cohen, Ruslan Salakhutdinov, and Christopher~D Manning.
\newblock Hotpotqa: A dataset for diverse, explainable multi-hop question answering.
\newblock {\em arXiv preprint arXiv:1809.09600}, 2018.

\bibitem{lightmem}
Jizhan Fang, Xinle Deng, Haoming Xu, Ziyan Jiang, Yuqi Tang, Ziwen Xu, Shumin Deng, Yunzhi Yao, Mengru Wang, Shuofei Qiao, et~al.
\newblock Lightmem: Lightweight and efficient memory-augmented generation.
\newblock {\em arXiv preprint arXiv:2510.18866}, 2025.

\bibitem{bai2023qwen}
Jinze Bai, Shuai Bai, Yunfei Chu, Zeyu Cui, Kai Dang, Xiaodong Deng, Yang Fan, Wenbin Ge, Yu~Han, Fei Huang, et~al.
\newblock Qwen technical report.
\newblock {\em arXiv preprint arXiv:2309.16609}, 2023.

\bibitem{chen2024bge}
Jianlv Chen, Shitao Xiao, Peitian Zhang, Kun Luo, Defu Lian, and Zheng Liu.
\newblock Bge m3-embedding: Multi-lingual, multi-functionality, multi-granularity text embeddings through self-knowledge distillation.
\newblock {\em arXiv preprint arXiv:2402.03216}, 2024.

\end{thebibliography}
